\documentclass[letterpaper]{article} % DO NOT CHANGE THIS
\usepackage{aaai2027}  % DO NOT CHANGE THIS
\usepackage[hyphens]{url}  % DO NOT CHANGE THIS
\usepackage{graphicx} % DO NOT CHANGE THIS
\usepackage{natbib}  % DO NOT CHANGE THIS AND DO NOT ADD ANY OPTIONS TO IT
\usepackage{caption} % DO NOT CHANGE THIS AND DO NOT ADD ANY OPTIONS TO IT
\usepackage{algorithm}
\usepackage{algorithmic}
\usepackage{booktabs}
\usepackage{amssymb} % for \checkmark
\usepackage{threeparttable}
\usepackage{amsmath}
\usepackage{multirow}
\usepackage{makecell}
\usepackage{graphicx}
\usepackage{newfloat}
\usepackage{listings}
\DeclareCaptionStyle{ruled}{labelfont=normalfont,labelsep=colon,strut=off} % DO NOT CHANGE THIS
\floatstyle{ruled}
\newfloat{listing}{tb}{lst}{}
\floatname{listing}{Listing}

\usepackage{listings}
\usepackage{xcolor}
\usepackage[most]{tcolorbox}

\lstdefinestyle{promptstyle}{
    basicstyle=\ttfamily\footnotesize,
    breaklines=true,
    breakatwhitespace=false,
    columns=fullflexible,
    frame=none,
    keywordstyle=\color{black},
    showstringspaces=false,
    xleftmargin=0pt,
    xrightmargin=0pt
}

\newtcolorbox{promptbox}[1]{
    colback=gray!5,
    colframe=gray!40,
    boxrule=0.4pt,
    arc=1pt,
    left=4pt, right=4pt, top=2pt, bottom=2pt,
    title=#1,
    fonttitle=\bfseries\small,
    coltitle=black,
    colbacktitle=gray!15
}

\usepackage{booktabs}

\title{Reward-Oracle MCTS for Formal Theorem Proving: Sample-Efficient Search and the Need for Kernel-Level Proof Auditing}

\author{
    Bodla Krishna Vamshi,
    Haizhao Yang\corresponding
}
\affiliations{
    University of Maryland, College Park\\
    kbodla@umd.edu, hzyang@umd.edu
}

\copyrighttext{Preprint. Under review.}
\begin{document}

\maketitle

\begin{abstract}
Formal theorem proving with large language models remains challenging due to the difficulty of navigating large proof search spaces efficiently. Existing tree search approaches either feed verbose compiler error messages directly into the generation context, increasing context usage during search, or employ non-standard evaluation protocols that prevent direct comparison with established baselines. We propose a three-role Monte Carlo Tree Search (MCTS) framework that treats the Lean 4 compiler purely as a reward oracle using compiler output as a scalar signal for UCB-guided tree updates without feeding error content into the generation context. Our framework decomposes proof search into three roles: a generator for proof attempts, a decomposer for subgoal decomposition, and a critic for subgoal quality evaluation. We evaluate across 4 benchmarks spanning competition mathematics and physics (MiniF2F, PutnamBench, LeanPhysBench, PhysLeandata) with three prover models at standard proof attempt budgets (PAB@16 to PAB@256). Our method achieves 87.1\% on MiniF2F with Goedel-Prover-V2-8B at PAB@256 and solves 26/659 PutnamBench problems at PAB@32 surpassing base sampling 18/659 at same proof attempt budget. Through an exhaustive axiom-level audit of every compiled proof, we further identify reward hacking in search-based theorem proving: DeepSeek-Prover-V2-7B produces proofs on PutnamBench that pass compilation and the standard sorry-token scan while depending on sorryAx. The audit removes 4 and 8 such proofs from whole-proof sampling at PAB@32 and PAB@128, and 11 and 19 from MCTS. We do not attribute these counts to the search procedure; we report them to establish that kernel-level auditing is necessary for compiler-verified evaluation.
\end{abstract}


\section{Introduction}

Formal theorem proving with large language models has emerged as a challenging frontier in AI reasoning, requiring models to navigate exponentially large proof search spaces while satisfying the strict correctness requirements of interactive theorem provers such as Lean 4 \cite{10.1007/978-3-030-79876-5_37} and Isabelle \cite{Nipkow2002}. Recent advances in specialized prover models \cite{lin2025goedelproverv2scalingformaltheorem, ren2025deepseekproverv2advancingformalmathematical, wang2025kiminaproverpreviewlargeformal} have demonstrated promising results on established benchmarks, yet the fundamental challenge of efficiently exploring proof search spaces remains open.

% Two broad paradigms have emerged for improving proof search beyond flat sampling. The first relies on \textit{compiler feedback}: feeding verbose error messages from the theorem prover back into the language model context to guide correction \cite{baba2026proveragentagentbasedframework, thakur2024incontextlearningagentformal}. The second paradigm uses \textit{tree search} with scalar reward signals \cite{gui2025hypertreeplanningenhancingllm}.

% % but existing methods either employ non-standard evaluation budgets that prevent direct comparison with established baselines \cite{baba2026proveragentagentbasedframework}, or lack the structured role decomposition necessary to effectively balance exploration and exploitation across the proof search space.

We propose a three-role Monte Carlo Tree Search (MCTS) framework built around a distinct architectural principle: the Lean 4 compiler is used exclusively as a \textit{reward oracle}, rather than as a source of textual feedback for subsequent generations. Unlike compiler-guided approaches that append error messages or proof-state diagnostics to the model context, our method converts compilation outcomes into scalar rewards that are propagated through the search tree and used only for UCB-guided node selection. This separation between verification and generation prevents compiler feedback from accumulating with search depth, while still allowing formal verification signals to influence future exploration through value backpropagation. Consequently, the framework performs iterative, verifier-guided search over natural-language proof plans without requiring compiler-error-conditioned generation.

% This design choice is empirically motivated: in preliminary experiments incorporating error feedback, model context limits were consistently exhausted by depth 2-3, preventing the deeper exploration that distinguishes MCTS from flat sampling.

Our framework decomposes proof search into three distinct roles: a \textbf{generator} that produces complete proof attempts, a \textbf{decomposer} that breaks problems into subgoals with temperature-decayed sampling to balance exploration and exploitation, and a \textbf{critic} that evaluates subgoal quality to provide a richer reward signal than binary compiler output alone. At each MCTS iteration, the UCB policy traverses the current tree from the root to select a leaf node based on backpropagated compiler and critic rewards.

We restrict our primary comparison to whole-proof sampling on the same frozen checkpoint, rather than to methods such as BFS-Prover \cite{xin2025bfsproverscalablebestfirsttree} , HunyuanProver \cite{li2025hunyuanproverscalabledatasynthesis}, which retrain or fine-tune the underlying prover model. Because these methods change the checkpoint itself via retrained value/policy networks, additional RL, or adversarially generated training data any performance delta conflates search-procedure gains with model gains, and cannot isolate the contribution of the search algorithm. Whole-proof sampling provides the cleanest primary baseline for isolating the effect of inference-time search while holding the prover checkpoint fixed. We additionally compare with inference-time structured-search baselines that use the same frozen checkpoints, prompts, and evaluation environment. Throughout the paper, we use the terms proof attempt budget and budget interchangeably, and denote it by $PAB@$.

% Empirical analysis reveals that successful solutions cluster at search depths 1--3. Through repeated root-to-leaf traversal, the UCB policy balances the exploitation of proof strategies with high empirical rewards against the exploration of less-visited alternatives. Each root-level child represents a distinct high-level proof approach, while backpropagated compiler and critic rewards progressively refine the allocation of search effort across these strategies. This emergent behavior, combined with temperature decay that increasingly exploits promising subtrees at lower temperatures, explains the performance gap between our full method and the no-decay ablation.

We evaluate across four benchmarks spanning competition mathematics and theoretical physics MiniF2F \cite{zheng2022minif2fcrosssystembenchmarkformal}, PutnamBench \cite{tsoukalas2024putnambenchevaluatingneuraltheoremprovers}, LeanPhysBench \cite{li2025lean4physicscomprehensivereasoningframework}, and PhysLeandata \cite{zhang2026physproveradvancingautomatictheorem} using three specialized prover models (Goedel-Prover-V2-8B \cite{lin2025goedelproverv2scalingformaltheorem}, DeepSeek-Prover-V2-7B \cite{ren2025deepseekproverv2advancingformalmathematical}, Kimina-Prover-Preview-Distill-7B \cite{wang2025kiminaproverpreviewlargeformal}) at standard proof attempt budgets (PAB@16 to PAB@256). On PutnamBench, our method solves 26/659 problems with Goedel-Prover-V2-8B at PAB@32 surpassing base sampling's 18/659 at same proof attempt budget. On MiniF2F, our method achieves 87.1\% with Goedel-Prover-V2-8B at PAB@256 and continues to improve consistently as the proof attempt budget increases.

Beyond strong empirical results, we identify and characterize a reward-hacking
phenomenon. DeepSeek-Prover-V2-7B, which was
previously reported to exploit a Lean 4.9.0 interface vulnerability
\cite{ren2025deepseekproverv2advancingformalmathematical}, produces
exploit-dependent successful proofs under both whole-proof sampling and MCTS
on PutnamBench. At PAB@32, the audit removes 4 successful proofs from
whole-proof sampling and 11 from MCTS; at PAB@128, it removes 8 and 19,
respectively. This finding highlights a fundamental
evaluation risk, making kernel-level proof auditing essential for search-based theorem-proving
evaluation. We further conduct a systematic multi-agent analysis showing that,
for the evaluated models, selectively using a heterogeneous critic can improve
proof search, whereas heterogeneous decomposition consistently degrades
performance. 

\section{Related work}

\subsection{Formal Theorem Proving with Large Language Models}
Early approaches to neural theorem proving focused on training specialized models to generate complete proof attempts in a single forward pass \cite{first2023baldurwholeproofgenerationrepair}. GPT-f \cite{polu2020generativelanguagemodelingautomated} demonstrated that language models pretrained on mathematical text could generate useful proof steps when fine-tuned on proof corpora, establishing the foundation for subsequent work. ReProver \cite{yang2023leandojotheoremprovingretrievalaugmented} introduced retrieval-augmented generation for tactic prediction, improving performance by conditioning proof generation on relevant premises retrieved from Mathlib.

\subsection{Tree Search for Theorem Proving}

Tree search methods have a long history in formal reasoning \cite{dong2025stpselfplayllmtheorem, xin2025bfsproverscalablebestfirsttree,li2025hunyuanproverscalabledatasynthesis}. HyperTree Proof Search \cite{lample2022hypertreeproofsearchneural} applies Monte Carlo Tree Search (MCTS) to Lean proof search using a learned value function jointly trained with the proof model to guide node expansion. COPRA \cite{thakur2024incontextlearningagentformal} performs context-sensitive proof search by maintaining proof state across attempts and incorporating compiler feedback into subsequent generations. Draft-Sketch-Prove \cite{jiang2023draftsketchproveguiding} decomposes proofs into high-level sketches that are subsequently completed by a tactic model, anticipating the use of intermediate proof decomposition. ReProver \cite{yang2023leandojotheoremprovingretrievalaugmented} improves tactic prediction through retrieval-augmented generation, while Prover Agent \cite{baba2026proveragentagentbasedframework} performs iterative compiler-guided search within a tree-based framework.

\subsection{Monte Carlo Tree Search in Language Model Reasoning}
MCTS has been applied broadly to improve LLM reasoning beyond theorem proving. \citet{yao2023treethoughtsdeliberateproblem} introduced Tree of Thoughts, using LLM-guided search over reasoning steps with value estimation for intermediate states. \citet{hao2023reasoninglanguagemodelplanning} proposed Reasoning via Planning, applying MCTS with LLM-based world models for multi-step reasoning tasks. \citet{feng2024alphazeroliketreesearchguidelarge} extended AlphaZero-style self-play to mathematical reasoning, training value and policy networks jointly. 

% Our work differs from these approaches in three key ways: (1) we decompose the search into three specialized roles rather than using a single model for all functions, (2) we apply temperature decay tied to both iteration count and node depth to manage exploration-exploitation balance, and (3) we operate in a formal verification setting where compiler execution provides a precise and automatically checkable reward
% signal, subject to the integrity of the evaluation interface enabling a pure reward oracle design that is not available in natural language reasoning settings.

\subsection{Whole-Proof Generation} A complementary research direction focuses on whole-proof generation \cite{first2023baldurwholeproofgenerationrepair}, where the model generates a complete Lean proof script in a single pass, typically accompanied by an extended reasoning trajectory. Recent advances in this paradigm have been driven by two major approaches. The first employs expert-iteration frameworks, which iteratively incorporate successfully verified proofs into the training corpus to progressively improve the prover \cite{polu2022formalmathematicsstatementcurriculum, wu2021tacticzerolearningprovetheorems, wu2025internlm25stepproveradvancingautomatedtheorem, lin2025leanstarlearninginterleavethinking,dong2025stpselfplayllmtheorem,lin2025goedelproverfrontiermodelopensource,lin2025goedelproverv2scalingformaltheorem}. The second leverages reinforcement learning guided by proof-assistant verification signals, enabling models to refine proof generation through verifier-based feedback \cite{kaliszyk2018reinforcementlearningtheoremproving,xin2024deepseekproverv15harnessingproofassistant,zhang2025leanabellproverposttrainingscalingformal,wang2025kiminaproverpreviewlargeformal,ren2025deepseekproverv2advancingformalmathematical,gloeckle2024abel,ji2025leanabellproverv2verifierintegratedreasoningformal,lin2025goedelproverv2scalingformaltheorem}.

% \subsection{Multi-Agent Frameworks for Reasoning}
% Multi-agent approaches to complex reasoning tasks have gained increasing attention. \citet{chan2023chatevalbetterllmbasedevaluators} demonstrated that multiple LLM agents debating can improve reasoning quality over single-model approaches. \citet{yuan2026marshalincentivizingmultiagentreasoning, wan2025remalearningmetathinkllms, sheng2023negotiatedreasoningprovablyaddressing} applied multi-agent reinforcement learning to collaborative problem solving, finding that role specialization improves performance on structured tasks. In the theorem proving context, \citet{zhao2024subgoalxlsubgoalbasedexpertlearning} proposed subgoal-based proof search where problems are decomposed into intermediate goals before individual tactics are generated — directly motivating our decomposer role.

% Our multi-agent analysis extends this line of work by systematically examining how distributional distance between models in different roles affects search quality, revealing that the critic and decomposer roles have opposite optimal assignment strategies — a finding not previously reported in the theorem proving literature.

\subsection{Reward Hacking in Reinforcement Learning}
Reward hacking where agents exploit unintended patterns in reward signals rather than solving the intended task is a well-documented failure mode in reinforcement learning \cite{wang2026rewardhackingeralarge}. In the LLM context, reward hacking has been observed in RLHF settings where models learn to exploit reward model weaknesses \cite{gao2022scalinglawsrewardmodel}. In formal theorem proving specifically, \citet{ren2025deepseekproverv2advancingformalmathematical} reported that DeepSeek-Prover-V2-7B exploited a Lean 4.9.0 compiler vulnerability where the \texttt{apply?} tactic failed to emit \texttt{sorry} declarations under certain conditions producing proofs that passed compilation without being mathematically valid.

\subsection{Lean 4 Proof Evaluation Infrastructure}
Automated evaluation of Lean 4 proofs requires interactive theorem prover infrastructure that can compile tactic-level proof attempts and return verification results programmatically. Several tools have been developed for this purpose, each with different design tradeoffs.

LeanDojo \cite{yang2023leandojotheoremprovingretrievalaugmented} is an open-source framework that provides a programmatic interface to Lean 4 through a REPL-based interaction model, enabling extraction of proof states, tactic execution, and premise retrieval from Mathlib. LeanDojo has become the standard evaluation infrastructure for many theorem proving benchmarks including MiniF2F and PutnamBench, and its premise extraction capabilities underpin retrieval-augmented approaches such as ReProver \cite{yang2023leandojotheoremprovingretrievalaugmented}.

LeanInteract \cite{leaninteract} extends the REPL-based approach with improved session management and support for parallel proof evaluation, reducing the overhead of maintaining multiple concurrent Lean processes. PyPantograph \cite{pantograph} takes a different approach, providing a Python-native interface to the Lean 4 kernel through direct FFI bindings rather than a REPL subprocess. 

Kimina Lean Server \cite{santos2025kiminaleanservertechnical} is purpose-built for large-scale parallel proof evaluation in the context of reinforcement learning and tree search methods. Unlike REPL-based tools that maintain sequential interaction with a single Lean process, Kimina Lean Server manages a pool of persistent Lean REPL instances that handle concurrent proof compilation requests enabling high-throughput evaluation at the sampling budgets required by tree search methods. The server exposes an HTTP interface that accepts complete Lean 4 proof strings and returns compilation results including success status, error messages, and \texttt{sorry} usage detection. Our method uses Kimina Lean Server as the compiler reward oracle throughout all experiments.

\section{Comparison with Existing Inference-Time Search Methods}
\label{sec:appendix_comparison}

To clarify the algorithmic distinctions between our framework and prior
inference-time search methods for formal theorem proving, Table~\ref{tab:comparison}
summarizes the major design choices. Unlike methods that condition future
generation on compiler feedback or proof-state diagnostics, our approach treats
the Lean verifier exclusively as a \emph{reward oracle}: only the scalar reward
is propagated through MCTS, while compiler messages are never inserted into the
LLM context. Furthermore, our search operates over natural-language proof plans
rather than formal proof states, allowing the search policy to allocate compute
through UCB-guided exploration while remaining independent of compiler-specific
feedback.

\begin{table*}[t]
\centering
\small
\renewcommand{\arraystretch}{1.25}
\setlength{\tabcolsep}{5pt}
\caption{Architectural comparison between our framework and closely related
inference-time search methods. ``Compiler usage'' refers to how Lean verification
is incorporated into the search process.}
\label{tab:comparison}

\resizebox{\textwidth}{!}{%
\begin{tabular}{lcccccc}
\toprule
\textbf{Method}
&
\textbf{Search}
&
\textbf{Search}
&
\textbf{Generator Receives}
&
\textbf{LLM}
&
\textbf{Search}
&
\textbf{Reward}\\

&
\textbf{Representation}
&
\textbf{Strategy}
&
\textbf{Compiler Feedback}
&
\textbf{Critic}
&
\textbf{Allocation}
&
\textbf{Backprop.}
\\
\midrule

COPRA
&
Formal proof state
&
Sequential refinement
&
\checkmark\;Compiler text
&
--
&
Fixed
&
-- \\

Prover Agent
&
Proof attempts
&
Tree refinement
&
\checkmark\;Compiler text
&
--
&
Heuristic expansion
&
Method-specific \\

BFS+CG
&
Natural-language plans
&
Breadth-first
&
$\times$ (scalar evaluation only)
&
\checkmark
&
Breadth-first
&
-- \\

One-shot decomposition
&
Natural-language plans
&
Single expansion
&
$\times$ (scalar evaluation only)
&
\checkmark
&
Flat expansion
&
-- \\

\midrule

\textbf{Reward-Oracle MCTS (Ours)}
&
\textbf{Natural-language proof plans}
&
\textbf{UCB-guided MCTS}
&
\textbf{$\times$ Scalar reward only}
&
\checkmark
&
\textbf{Adaptive}
&
\checkmark
\\

\bottomrule
\end{tabular}}
\end{table*}

\section{Method}

\begin{figure*}[t]
\centering
\includegraphics[width=0.8\textwidth]{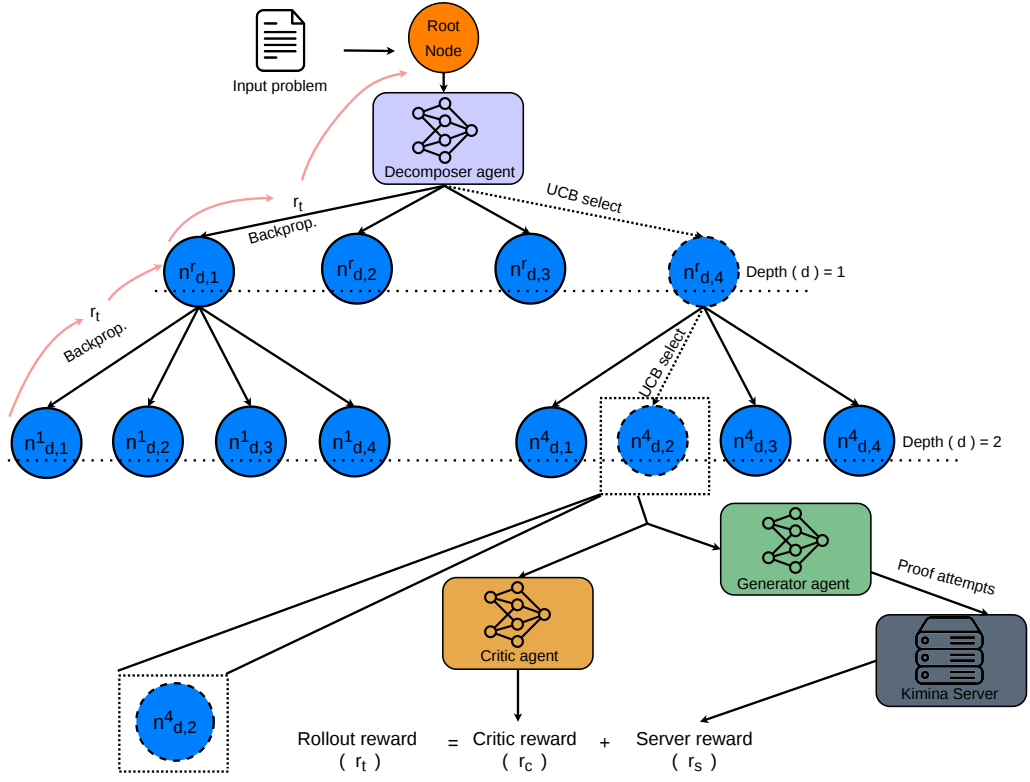} % Reduce the figure size so that it is slightly narrower than the column.
\caption{Overview of the proposed search framework. Each node $n^P_{d,N}$ represents the $k$-th child at depth $d$, with parent $P$:$P$ = $r$ denotes the root.}
\label{fig:mcts}
\end{figure*}

Our framework decomposes proof search into three specialized roles: a \textbf{generator}, a \textbf{decomposer}, and a \textbf{critic} coordinated by a Monte Carlo Tree Search procedure. Each node $n^P_{d,k}$ in the search tree represents a subgoal state at depth $d$, child index $k$, with parent node $P$ (where $P{=}r$ denotes the root). The tree is initialized with a single root node $n^r_{0,0}$ corresponding to the original theorem $T$. We describe the four phases of each MCTS iteration below, with the overall procedure illustrated in Figure~\ref{fig:mcts}. Throughout this paper, “states” in our method refer to natural-language proof-plan search states rather than formal Lean 4 proof states.

%-------------------------------------------------------------------
\subsection{Selection}
%-------------------------------------------------------------------

Beginning from the root node $n^r_{0,0}$, the selection phase traverses the tree by repeatedly choosing the child node that maximizes the UCB criterion:

\begin{equation}
\text{UCB}(n^P_{d,k}) \;=\; Q(n^P_{d,k}) \;+\; c\sqrt{\frac{\ln N\!\left(\text{parent}(n^P_{d,k})\right)}{N(n^P_{d,k})}}
\label{eq:ucb}
\end{equation}

\noindent where $Q(n)$ is the empirical mean rollout reward of node $n$, $N(n)$ is its visit count, and $c=\sqrt2$ is the exploration constant. For an unvisited node, the exploration term is treated as $+\infty$, ensuring that every newly generated
child is evaluated before previously visited children are reselected. Traversal continues until a leaf node is reached that is either terminal or has not yet been fully expanded. If the selected node is non-terminal and has not yet been expanded, it is expanded into $K$ child nodes. Previously expanded internal nodes remain eligible for future traversal, and their children are repeatedly selected according to the UCB policy as the search proceeds. Thus, the tree is incrementally refined over successive iterations as visit counts and value estimates are updated through backpropagation.
%-------------------------------------------------------------------
\subsection{Expansion}
%-------------------------------------------------------------------

The selected node $n$ at depth $d$ is expanded by generating $K{=}4$ child subgoals ($g_k$) using the decomposer role. The decomposer is prompted with the original theorem and the trajectory from the root to the current node, namely, the sequence of subgoals accumulated along that path. It proposes the next intermediate proof step as a natural language subgoal, producing $K$ diverse candidates that become child nodes $n^n_{d+1,1}, \ldots, n^n_{d+1,K}$ in the tree.

A key design choice in expansion is \textbf{temperature decay}: the decomposer sampling temperature $\tau_d$ is annealed according to both the current depth $d$ and the iteration index $i$ as:

\begin{equation}
\tau_d \;=\; \tau_0 \cdot e^{-0.12\,d} \cdot e^{-0.03\ln(1+i)}
\label{eq:temp}
\end{equation}

\noindent starting from $\tau_0{=}0.7$. At shallow depths and early iterations, the higher temperature encourages diverse subgoal proposals, enabling broad exploration of distinct proof strategies. At greater depths and later iterations, the lower temperature concentrates sampling on more focused, committed proof directions. This schedule directly governs the exploration--exploitation balance of the search. Our ablation analysis, provided in the supplementary appendix submitted with the paper, shows that removing temperature decay consistently degrades performance across the evaluated models, indicating that it is a critical architectural component of our framework.

%-------------------------------------------------------------------
\subsection{Evaluation}
%-------------------------------------------------------------------

Each newly expanded child node $n^n_{d+1,k}$ is evaluated using two complementary reward signals that together form the rollout reward $r_t = r_c + r_s$.

\paragraph{Critic reward $r_c \in [0,1]$.}
The critic role assesses the quality of the proposed subgoal $g_k$ by sampling the critic LLM five times at a fixed low temperature of $\tau{=}0.3$ and averaging the scores. The low temperature ensures stable, consistent evaluation rather than diverse sampling. The critic provides a soft quality signal that compensates for the binary nature of compiler feedback a subgoal that does not immediately yield a compiling proof may still receive high critic reward if it represents a promising intermediate step toward the final proof.  The critic outputs a score \(\hat{r}_c \in [0, 100]\), which we normalize as \(r_c = \hat{r}_c / 100 \in [0, 1]\). 

% The server reward \(r_s \in \{0,1/S,2/S \ldots, 1\}\) is the number of successful proof attempts among the \(S\) generator samples.

% We retain the unnormalized count rather than using a success proportion so that nodes producing multiple valid proofs receive stronger empirical evidence during UCB updates. 

% Consequently, the critic has greater relative influence when \(S\) is small and compiler evidence is sparse, while its influence decreases as the number of proof attempts per node increases.

\paragraph{Server reward $r_s \in [0,1]$.}
The generator role produces $S$ complete Lean~4 proof attempts for each child node, conditioned on the full trajectory from root to that node at a fixed temperature of $\tau{=}0.7$. Each attempt is submitted to the Kimina Lean server \cite{santos2025kiminaleanservertechnical}, which compiles the proof and checks for the absence of \texttt{sorry} declarations. Let $s_k$ denote the number of attempts that compile successfully. We define the normalized server reward as $r_s = s_k / S$, \(r_s \in \{0,1/S,2/S \ldots, 1\}\). Crucially, we use compiler output solely as a scalar count signal no error message content is fed back into the generation context. 

% \noindent If any generated proof passes compilation and is \texttt{sorry}-free, it is immediately returned as the solution without completing the remaining iterations. Otherwise the combined rollout reward $r_t = r_c + r_s$ is recorded for each child and used in backpropagation.

%-------------------------------------------------------------------
\subsection{Backpropagation}
%-------------------------------------------------------------------

After rollout evaluation, the obtained rollout reward $r_t$ is propagated from the expanded node back to the root along the selected search path. For every node on this path, the visit count and cumulative reward are updated as
\[
N(n) \leftarrow N(n)+1,
\qquad
W(n) \leftarrow W(n)+r_t,
\]
and the node value used by UCB is computed as
\[
Q(n)=\frac{W(n)}{N(n)}.
\]
These updates follow the standard MCTS backpropagation procedure. These updated statistics are subsequently used by the UCB selection policy during future iterations to balance exploration and exploitation.

After backpropagation, the next iteration begins with a fresh selection phase from the root. The updated visit counts and value estimates allow the UCB policy to allocate more search effort to promising subtrees while continuing to explore less-visited alternatives. 

% Empirically, solutions cluster at depths 1--3, indicating that the search frequently finds successful proof strategies after only a small number of decomposition steps.

%-------------------------------------------------------------------
\subsection{Proof attempt Budget}
%-------------------------------------------------------------------

At each of $N$ MCTS iterations, one leaf node is selected according to the UCB policy and expanded into $K$ child nodes. Each child is evaluated using $S$ complete proof attempts, resulting in $K \times S$ generator rollouts per iteration. Therefore, the total proof-attempt sampling budget is exactly $N \times K \times S$ per problem. We scale the proof attempt budget by varying $S$ while keeping $N$ and $K$ fixed:

\begin{equation}
\text{PAB@}B \;=\; N \times K \times S 
\;=\; 4 \times 4 \times S
\label{eq:budget}
\end{equation}

\noindent giving $S{=}2$ for PAB@32, $S{=}4$ for PAB@64, $S{=}8$ for PAB@128, and $S{=}16$ for PAB@256. Throughout this paper, PAB@B denotes proof-attempt budget.

\begin{table*}[t]
\centering
\small
\setlength{\tabcolsep}{3pt}
\begin{threeparttable}
\caption{Results on the miniF2F test set for different methods. All results are averaged over 5 random seeds.}
\label{tab:minif2f_results}
\begin{tabular}{@{}llccccc@{}}
\toprule
\textbf{Model} & \textbf{Method} & \textbf{Size} & \textbf{Generator} & \textbf{Decomposer} & \textbf{Critic} & \textbf{Success (\%)} \\
 & & & \textbf{completions} & \textbf{completions} & \textbf{completions} & \\
\midrule
\multirow{3}{*}{Kimina}
 & BFS+CG                 & \multirow{3}{*}{7B} & $4\times2\times4$  & $4\times4$  & $4\times4$  & $62.5 \pm 0.6$ \\
 & One-shot decomposition &                     & $16\times2\times1$ & $16\times1$ & $16\times1$ & $61.3 \pm 0.6$ \\
 & Agent (ours)           &                     & $4\times2\times4$  & $4\times4$  & $4\times4$  & $\mathbf{64.3 \pm 0.8}$ \\
\midrule
\multirow{3}{*}{DeepSeek}
 & BFS+CG                 & \multirow{3}{*}{7B} & $4\times2\times4$  & $4\times4$  & $4\times4$  & $75.4 \pm 0.6$ \\
 & One-shot decomposition &                     & $16\times2\times1$ & $16\times1$ & $16\times1$ & $74.5 \pm 0.3$ \\
 & Agent (ours)           &                     & $4\times2\times4$  & $4\times4$  & $4\times4$  & $\mathbf{77.1 \pm 0.3}$ \\
\midrule
\multirow{3}{*}{Goedel}
 & BFS+CG                 & \multirow{3}{*}{8B} & $4\times2\times4$  & $4\times4$  & $4\times4$  & $82.7 \pm 0.4$ \\
 & One-shot decomposition &                     & $16\times2\times1$ & $16\times1$ & $16\times1$ & $81.7 \pm 0.4$ \\
 & Agent (ours)           &                     & $4\times2\times4$  & $4\times4$  & $4\times4$  & $\mathbf{84.2 \pm 0.5}$ \\
\bottomrule
\end{tabular}
\begin{tablenotes}
\footnotesize
\item BFS indicates breadth first search, CG indicates critic guided search
\end{tablenotes}
\end{threeparttable}
\end{table*}

\begin{table}[t]
\centering
\small
\setlength{\tabcolsep}{3pt}
\caption{Results on miniF2F test set for different proof attempt budgets and methods. All results are averaged over 5 random seeds. Prover Agent is reproduced under our evaluation setting at a proof attempt budget of 260, since only at this budget does its full iterative-refinement pipeline execute.}
\label{tab:minif2f_results_main}
\begin{tabular}{@{}llccc@{}}
\toprule
\textbf{Model} & \textbf{Method} & \textbf{Size} & \textbf{Budget} & \textbf{Success (\%)} \\

% \midrule

% STP & Whole proof & 7B & $25600$ & 67.6 \\
% Hypertree & Tree search & 0.7B & $64$ & 41.0 \\
% BFSProver & Tree search & 7B &  $2048$ & 70.83  \\
% Hunyanproverv16 & Tree search & 7B & $600$ & 68.4 \\
% COPRA  & Agent & - & $1 $ & 26.63 \\

\midrule

\multirow{6}{*}{Kimina}
 & \multirow{2}{*}{Whole-proof} & \multirow{6}{*}{7B} & 32 & $62.8 \pm 0.8$ \\
 & & & 64 & $65.1 \pm 0.3$ \\
% \cmidrule(l){2-2} \cmidrule(l){4-5}
%  & \multirow{2}{*}{Agent (critic only)} & & 32 & $61.4 \pm 0.9$ \\
%  & & & 64 & $62.7 \pm 0.6$ \\
\cmidrule(l){2-2} \cmidrule(l){4-5}
 & \multirow{2}{*}{Agent (ours)} & & 32 & $64.3 \pm 0.8$ \\
 & & & 64 & $\mathbf{66.9 \pm 0.2}$ \\
\midrule
\multirow{7}{*}{DeepSeek}
 & \multirow{2}{*}{Whole-proof} & \multirow{7}{*}{7B} & 32 & $75.2 \pm 0.5$ \\
 & & & 64 & $76.3 \pm 0.7$ \\
  & & & 128 & $77.5 \pm 0.3$ \\
  & & & 256 & $78.2 \pm 0.1$ \\
% \cmidrule(l){2-2} \cmidrule(l){4-5}
%  & \multirow{2}{*}{Agent (critic only)} & & 32 & $73.7 \pm 0.7$ \\
%  & & & 64 & $74.2 \pm 0.3$ \\
\cmidrule(l){2-2} \cmidrule(l){4-5}
 & Prover Agent & & 260 & $81.7 \pm 0.3$ \\
\cmidrule(l){2-2} \cmidrule(l){4-5}

 & \multirow{3}{*}{Agent (ours)} & & 32 & $77.1 \pm 0.3$ \\
 & & & 64 & $79.3 \pm 0.4$ \\
 & & & 128 & $81.4 \pm 0.2$ \\
  & & & 256 & $\mathbf{82.6 \pm 0.1}$ \\
\midrule
\multirow{8}{*}{Goedel}
 & \multirow{2}{*}{Whole-proof} & \multirow{8}{*}{8B} & 32 & $82.4 \pm 0.6$ \\
 & & & 64 & $83.3 \pm 0.4$ \\
 & & & 128 & $83.9 \pm 0.2$ \\
  & & & 256 & $84.7 \pm 0.1$ \\
% \cmidrule(l){2-2} \cmidrule(l){4-5}
%  & \multirow{2}{*}{Agent (critic only)} & & 32 & $81.6 \pm 0.6$ \\
%  & & & 64 & $82.5 \pm 0.5$ \\
\cmidrule(l){2-2} \cmidrule(l){4-5}
 & Prover Agent & & 260 & $86.2 \pm 0.1$ \\
\cmidrule(l){2-2} \cmidrule(l){4-5}

 & \multirow{4}{*}{Agent (ours)} & & 32 & $84.2 \pm 0.5$ \\
 & & & 64 & $85.6 \pm 0.4$ \\
 & & & 128 & $86.3 \pm 0.2$ \\
 & & & 256 & $\mathbf{87.1 \pm 0.2}$ \\
\bottomrule
\end{tabular}
% \begin{tablenotes}
% \footnotesize
% \item Prover Agent is reproduced under our evaluation setting at a sampling budget of 260, since only at this budget does its full iterative-refinement pipeline execute.
% \end{tablenotes}
\end{table}

\begin{table}[t]
\centering
\small
\setlength{\tabcolsep}{3pt}
\caption{Results on miniF2F test set for reward ablations. All results are averaged over 5 random seeds.}
\label{tab:reward_ablation}
\begin{tabular}{@{}llccc@{}}
\toprule
\textbf{Model} & \textbf{Method} & \textbf{Size} & \textbf{PAB} & \textbf{Success (\%)} \\
\midrule
\multirow{4}{*}{DeepSeek}
 & \multirow{2}{*}{Agent (critic only)} & \multirow{4}{*}{7B} & 32 & $73.7 \pm 0.7$ \\
 & & & 64 & $74.2 \pm 0.3$ \\
\cmidrule(l){2-2} \cmidrule(l){4-5}
 & \multirow{4}{*}{Agent (critic+reward)} & & 32 & $77.1 \pm 0.3$ \\
 & & & 64 & $79.3 \pm 0.4$ \\
 & & & 128 & $81.4 \pm 0.2$ \\
  & & & 256 & $82.6 \pm 0.1$ \\
\midrule
\multirow{6}{*}{Goedel}
 & \multirow{2}{*}{Agent (critic only)} & \multirow{6}{*}{8B} & 32 & $81.6 \pm 0.6$ \\
 & & & 64 & $82.5 \pm 0.5$ \\
\cmidrule(l){2-2} \cmidrule(l){4-5}
 & \multirow{4}{*}{Agent (critic+reward)} & & 32 & $84.2 \pm 0.5$ \\
 & & & 64 & $85.6 \pm 0.4$ \\
 & & & 128 & $86.3 \pm 0.2$ \\
 & & & 256 & $\mathbf{87.1 \pm 0.2}$ \\
\bottomrule
\end{tabular}
\end{table}

\begin{table}[t]
\centering
\small
\setlength{\tabcolsep}{3pt}
\caption{Results on PutnamBench for different methods and proof attempt budgets. Following prior PutnamBench evaluations, results are reported as a single benchmark run over all 659 problems}
\label{tab:putnam_results}
\begin{tabular}{@{}llccc@{}}
\toprule
\textbf{Model} & \textbf{Method} & \textbf{Size} & \textbf{PAB} & \textbf{\#Solved} \\
\midrule
\multirow{2}{*}{DeepSeek}
 & \multirow{2}{*}{Whole-proof} & \multirow{2}{*}{7B} & 32 & 9/659 \\
  & & & 32$^{*}$ & $\mathbf{13/659}$ \\
 & & & 128 & 10/659 \\
  & & & 128$^{*}$ & $\mathbf{18/659}$ \\
\cmidrule(l){2-2} \cmidrule(l){4-5}
 & \multirow{4}{*}{Agent (ours)} & & 32 & 16/659 \\
 & & & 32$^{*}$ & $\mathbf{27/659}$ \\
 & & & 128 & 25/659 \\
 & & & 128$^{*}$ & $\mathbf{44/659}$ \\
\midrule
\multirow{6}{*}{Goedel}
 & \multirow{2}{*}{Whole-proof} & \multirow{6}{*}{8B} & 32 & 18/659 \\
 & & & 128 & 22/659 \\
\cmidrule(l){2-2} \cmidrule(l){4-5}
 & \multirow{2}{*}{Agent (ours)} & & 32 & 26/659 \\
 & & & 128 & $36/659$ \\
% \cmidrule(l){2-2} \cmidrule(l){4-5}
%  & Prover Agent & & 110 & 25/659 \\
\bottomrule
\end{tabular}
\vspace{2pt}
\\
{\footnotesize $^{*}$Potential exploit.}
% \begin{tablenotes}
% \footnotesize
% \item Prover Agent is reproduced under our evaluation setting.
% \end{tablenotes}
\end{table}

\begin{table}[t]
\centering
\small
\setlength{\tabcolsep}{3pt}
\caption{Results on PhysLeandata test set. Results are averaged across 5 random seeds.}
\label{tab:pylean_results}
\begin{tabular}{@{}llccc@{}}
\toprule
\textbf{Model} & \textbf{Method} & \textbf{Size} & \textbf{PAB} & \textbf{Success (\%)} \\
\midrule
\multirow{2}{*}{Kimina} & Whole-proof & \multirow{2}{*}{7B} & 16 & $25.2 \pm 0.6$ \\
 & Agent (critic+reward) & & 16 & $\mathbf{27.5 \pm 0.6}$ \\
\midrule
\multirow{2}{*}{DeepSeek} & Whole-proof & \multirow{2}{*}{7B} & 16 & $33.8 \pm 0.5$ \\
 & Agent (critic+reward) & & 16 & $\mathbf{36.2 \pm 0.6}$ \\
\midrule
\multirow{2}{*}{Goedel} & Whole-proof & \multirow{2}{*}{8B} & 16 & $31.9 \pm 0.7$ \\
 & Agent (critic+reward) & & 16 & $\mathbf{33.8 \pm 0.4}$ \\
\bottomrule
\end{tabular}
\end{table}

\begin{table}[t]
\centering
\small
\setlength{\tabcolsep}{4pt}
\caption{Results on LeanPhysBench (proof attempt budget (PAB@B) = 16). Results are averaged across 5 random seeds. The PhysLib library is directly passed as context. Critic+server rewards used for all our method configurations.}
\label{tab:leanphysbench_results}
\begin{tabular}{@{}llc@{}}
\toprule
\textbf{Model} & \textbf{Method} & \textbf{Success (\%)} \\
\midrule
\multirow{4}{*}{Kimina}
 & Whole-proof & $9.3 \pm 0.3$ \\
 & Whole-proof (w/ PhysLib) & $13.5 \pm 0.7$ \\
 & Agent (ours, no PhysLib) & $11.5 \pm 0.6$ \\
 & Agent (ours, w/ PhysLib) & $\mathbf{15.0 \pm 0.7}$ \\
\midrule
\multirow{4}{*}{DeepSeek}
 & Whole-proof & $11.5 \pm 0.2$ \\
 & Whole-proof (w/ PhysLib) & $14.5 \pm 0.2$ \\
 & Agent (ours, no PhysLib) & $13.0 \pm 0.4$ \\
 & Agent (ours, w/ PhysLib) & $\mathbf{16.5 \pm 0.6}$ \\
\midrule
\multirow{4}{*}{Goedel}
 & Whole-proof & $10.0 \pm 0.2$ \\
 & Whole-proof (w/ PhysLib) & $15.0 \pm 0.2$ \\
 & Agent (ours, no PhysLib) & $12.0 \pm 0.4$ \\
 & Agent (ours, w/ PhysLib) & $\mathbf{16.5 \pm 0.3}$ \\
\bottomrule
\end{tabular}
\end{table}

% \begin{table}[t]
% \centering
% \small
% \setlength{\tabcolsep}{4pt}
% \caption{Depth-wise breakdown of solved/attempted problems by category  for Goedel-Prover-V2-8B at PAB@256 on the MiniF2F test set. Each entry  $a/b$ denotes $a$ problems solved out of $b$ problems that reached this depth during search.}
% \label{tab:depth_wise_results}
% \begin{tabular}{@{}lcccc@{}}
% \toprule
% \textbf{Category} & \textbf{D1} & \textbf{D2} & \textbf{D3} & \textbf{D4} \\
% \midrule
% AIME           & 8/8   & 1/3 & 0/1 & 2/3 \\
% Algebra        & 78/78 & 2/4 & 0/3 & 0/5 \\
% AMC            & 39/39 & 4/4 & 2/2 & --  \\
% IMO            & 8/8   & 1/2 & 1/2 & 2/8 \\
% Induction      & 5/5   & --  & 0/2 & 0/1 \\
% Number Theory  & 54/54 & 3/5 & 3/5 & 0/4 \\
% \bottomrule
% \end{tabular}
% \begin{tablenotes}
% \small
% \item[--] $(--)$ denotes no problems of this category 
% reached the given search depth.
% \end{tablenotes}
% \end{table}

\section{Results}
\label{sec:results}

We evaluate our framework across four benchmarks MiniF2F \cite{zheng2022minif2fcrosssystembenchmarkformal}, 
PutnamBench \cite{tsoukalas2024putnambenchevaluatingneuraltheoremprovers}, PhysLeandata \cite{zhang2026physproveradvancingautomatictheorem}, and LeanPhysBench 
\cite{li2025lean4physicscomprehensivereasoningframework} using three specialized prover models: 
Goedel-Prover-V2-8B \cite{lin2025goedelproverv2scalingformaltheorem}, DeepSeek-Prover-V2-7B \cite{ren2025deepseekproverv2advancingformalmathematical}, 
and Kimina-Prover-Preview-Distill-7B \cite{wang2025kiminaproverpreviewlargeformal}. All results are averaged over 5 random 
seeds. Unlike much prior work in this area, we report mean $\pm$ standard deviation across 5 random seeds for all experiments to support statistical reliability of the reported gains.

\paragraph{Baseline description} \textbf{BFS+CG}. Breadth-first expansion of the decomposition tree in which the critic score alone ranks and selects nodes for expansion at each level, with no UCB criterion and no backpropagation of compiler reward along the search path.

\textbf{One-shot decomposition}. All 16 subgoal candidates are generated in a single flat expansion from the root and each is evaluated independently, with no iterative reselection isolating whether gains derive from decomposition scaffolding alone rather than from iterative, reward-guided search.

Both baselines use the identical frozen checkpoints for the generator, decomposer, and critic roles as our framework, and identical prompt templates and decoding configurations; they differ from our method only in the search mechanism that allocates the fixed proof-attempt budget.
\subsection{Main Results}
\label{sec:main}
%-------------------------------------------------------------------

Table~\ref{tab:minif2f_results} and ~\ref{tab:minif2f_results_main} reports results on MiniF2F test set. Our method
consistently outperforms whole proof sampling methods across all three models at matched proof attempt budgets. On Goedel-Prover-V2-8B our method achieves $\mathbf{84.2 \pm 0.5\%}$ at PAB@32, surpassing whole-proof baseline ($82.4\%$) (Table~\ref{tab:minif2f_results_main}). To analyze the importance of individual reward functions, we perform ablations comparing the critic-only reward against the combined critic + verifier reward (Table~\ref{tab:reward_ablation}). 

% An ablation isolating the verifier-only reward is not feasible, since in the initial stage, when all proof attempts are unsuccessful, all nodes become equiprobable for exploration.

To directly compare against a feedback-driven baseline on the same checkpoint, we reproduce Prover Agent \cite{baba2026proveragentagentbasedframework} using their released implementation with Goedel-Prover-V2-8B as the prover model, under our evaluation environment (Lean 4.15.0, Mathlib commit 9837ca9d). At their native sample budget of 260, we obtain $\mathbf{86.2 \pm 0.1\%}$ on MiniF2F, compared to $\mathbf{87.1\pm 0.2\%}$ for our method at PAB@256, despite Prover Agent's use of direct compiler-error feedback in context versus our oracle-only signal.

On PutnamBench (Table~\ref{tab:putnam_results}), our method solves $26/659$ problems with Goedel-Prover-V2-8B at PAB@32 and $36/659$ at PAB@128, compared to $18/659$ and $22/659$ for whole-proof sampling. On PhysLeandata and LeanPhysBench
(Tables~\ref{tab:pylean_results},~\ref{tab:leanphysbench_results}),
our method consistently outperforms whole-proof sampling across all
three models at the matched budget of PAB@16, with gains of $+1.9\%$
to $+2.4\%$ on PhysLeandata and $+1.5\%$ to $+2.0\%$ on LeanPhysBench
with PhysLib context. Notably, our method without PhysLib context
approaches the performance of whole-proof sampling with PhysLib
suggesting that structured search partially compensates for the absence
of domain-specific library context. Consistent gains across both physics
benchmarks demonstrate that our method generalizes beyond mathematical
domains where model familiarity is minimal.

\section{Inference-token efficiency}
In this section we additionally compare cumulative inference-token usage at PAB@32  (Table ~\ref{tab:role_token_breakdown}). All reported counts are averaged across five complete runs per sample. We use vLLM package for inference with one model instance loaded per GPU; the same instance serves the generator, decomposer, and critic roles. We use 4 Nvidia H200-SXM GPUs for inference. Multiple completions are generated within batched requests: whole-proof sampling produces 32 outputs from one request, while each critic evaluation produces five outputs in one request whose scores are subsequently averaged. To ensure strict budget matching, neither method terminates early after finding a valid proof.

Despite the additional decomposition and critic computations, our framework consistently consumes fewer inference tokens across all three models. Relative to whole-proof sampling, total token usage decreases by 32.29\% for Goedel-Prover-V2-8B, 33.17\% for DeepSeek-Prover-V2-7B, and 33.07\% for Kimina-Prover-7B. Output-token usage decreases by 35.35\%, 36.07\%, and 36.05\%, respectively. Averaged across the three models, our framework reduces output-token generation by 35.83\% and total token consumption by 32.84\%. These results show that the short decomposer and critic interactions substantially reduce the number of generator output tokens consumed under the same proof-attempt budget, more than offsetting their additional token overhead. Both methods use an identical vLLM deployment with one model instance per GPU and the same maximum GPU-memory-utilisation setting of 0.95.

\begin{table*}[t]
\centering
\small
\caption{Role-level inference-token usage at PAB@32, averaged
across five runs. Token counts are cumulative across all role
completions in one complete benchmark run.}
\label{tab:role_token_breakdown}
\begin{tabular}{llrrr}
\toprule
\textbf{Model} & \textbf{Role/Method} &
\textbf{Input} & \textbf{Output} & \textbf{Total} \\
\midrule
\multirow{4}{*}{Goedel-Prover-V2-8B}
 & Decomposer  & 1,438 & 1,449   & 2,887 \\
 & Critic      & 4,126 & 196     & 4,322 \\
 & Generator   & 4,231 & 203,580 & 207,811 \\
 & Whole-proof & 117   & 317,420 & 317,537 \\
\midrule
\multirow{4}{*}{DeepSeek-Prover-V2-7B}
 & Decomposer  & 1,347 & 1,392   & 2,739 \\
 & Critic      & 4,021 & 187     & 4,208 \\
 & Generator   & 4,058 & 204,113 & 208,171 \\
 & Whole-proof & 117   & 321,762 & 321,879 \\
\midrule
\multirow{4}{*}{Kimina-Prover-7B}
 & Decomposer  & 1,427 & 1,466   & 2,893 \\
 & Critic      & 4,083 & 185     & 4,268 \\
 & Generator   & 4,135 & 203,611 & 207,746 \\
 & Whole-proof & 117   & 320,986 & 321,103 \\
\bottomrule
\end{tabular}
\end{table*}

%-------------------------------------------------------------------
\subsection{Effect of Adaptive Decomposer Temperature}
\label{sec:ablation}
%-------------------------------------------------------------------

Table~\ref{tab:temp_variation_results} reports the effect of removing 
temperature decay from the decomposer, replacing the annealed schedule 
\begin{equation}
\tau_d \;=\; \tau_0 \cdot e^{-0.12\,d} \cdot e^{-0.03\ln(1+i)}
\label{eq:temp}
\end{equation}

(Equation~\ref{eq:temp}) with a fixed temperature of $\tau{=}0.7$ throughout. 
Removing temperature decay consistently degrades performance across all three 
models: Kimina-Prover-7B drops from $64.3 \pm 0.8\%$ to $62.5 \pm 0.5\%$ 
($-1.8\%$), DeepSeek-Prover-V2-7B from $77.1 \pm 0.3\%$ to $74.6 \pm 0.3\%$ 
($-2.5\%$), and Goedel-Prover-V2-8B from $84.2 \pm 0.5\%$ to $81.9 \pm 0.3\%$ 
($-2.3\%$).

These consistent drops across all models confirm that temperature decay is 
a critical architectural component rather than a model-specific artifact. 
Without decay, the decomposer maintains high temperature throughout the search 
continuing to propose diverse subgoals even at deeper depths and later 
iterations where the search should be committing to promising proof directions. 
This prevents the tree from effectively exploiting high-reward subtrees 
identified by backpropagated compiler rewards, reducing the search to 
essentially diverse but unguided sampling.

\begin{table}[t]
\centering
\small
\setlength{\tabcolsep}{4pt}
\caption{Results with and without temperature variation on miniF2F test set (proof attempt budget = 32; critic + server rewards used for all configurations). Results are averaged across 5 random seeds.}
\label{tab:temp_variation_results}
\begin{tabular}{@{}llc@{}}
\toprule
\textbf{Model} & \textbf{Method} & \textbf{Success (\%)} \\
\midrule
\multirow{2}{*}{Kimina}
 & Agent (ours, w/ temp.\ variation) & $\mathbf{64.3 \pm 0.8}$ \\
 & Agent (ours, w/o temp.\ variation) & $62.5 \pm 0.5$ \\
\midrule
\multirow{2}{*}{DeepSeek}
 & Agent (ours, w/ temp.\ variation) & $\mathbf{77.1 \pm 0.3}$ \\
 & Agent (ours, w/o temp.\ variation) & $74.6 \pm 0.3$ \\
\midrule
\multirow{2}{*}{Goedel}
 & Agent (ours, w/ temp.\ variation) & $\mathbf{84.2 \pm 0.5}$ \\
 & Agent (ours, w/o temp.\ variation) & $81.9 \pm 0.3$ \\
\bottomrule
\end{tabular}
\end{table}

\subsection{Multi-Agent Role Analysis}
\label{sec:multiagent}
%-------------------------------------------------------------------

Table~\ref{tab:multi_agent_results} reports results from our systematic
multi-agent analysis on Minif2f test set for proof attempt budget of 32 (PAB@32), where Goedel-Prover-V2-8B is fixed as the generator and all combinations of Goedel-Prover-V2-8B, DeepSeek-Prover-V2-7B, and Kimina-Prover-7B are evaluated in the decomposer and critic roles. The objective is to examine whether introducing heterogeneous models
into supporting roles can improve proof search performance over the homogeneous baseline.

Among the evaluated critic assignments, heterogeneous critics exhibit mixed behavior. Using DeepSeek-Prover-V2-7B as the critic achieves the best overall performance ($\mathbf{86.1 \pm 0.3\%}$), improving upon the homogeneous configuration ($84.2 \pm 0.5\%$) by $1.9$ percentage points. In contrast, replacing the critic with Kimina-Prover-7B decreases performance to $82.7 \pm 0.4\%$. These results indicate that, for the models evaluated here, an appropriately chosen heterogeneous critic can provide complementary evaluation signals that improve search quality, although this benefit is not observed for every model substitution.

The decomposer role exhibits a different trend. Replacing the homogeneous Goedel-Prover-V2-8B decomposer with either DeepSeek-Prover-V2-7B or Kimina-Prover-7B consistently reduces performance, achieving $83.8 \pm 0.4\%$ ($-0.4$ points) and
$82.2 \pm 0.5\%$ ($-2.0$ points), respectively. Furthermore, combining heterogeneous models in both supporting roles does not preserve the critic improvement. For example, using DeepSeek-Prover-V2-7B as both decomposer and critic achieves only $80.9 \pm 0.3\%$, substantially below both the homogeneous baseline ($84.2 \pm 0.5\%$) and the configuration that changes only the critic ($86.1 \pm 0.3\%$). This
suggests that, for the evaluated models, the performance gain from a heterogeneous critic depends on maintaining a homogeneous generator--decomposer pair.

Overall, within the scope of the evaluated models and benchmark, the strongest configuration pairs a Goedel-Prover-V2-8B generator and decomposer with a DeepSeek-Prover-V2-7B critic. These findings suggest that selectively introducing heterogeneity into the critic role is a promising direction for improving theorem-proving search, while additional experiments across a broader range of generator models and benchmarks are needed to determine the extent to which these observations
generalize.

\begin{table}[t]
\centering
\small
\setlength{\tabcolsep}{4pt}
\caption{Multi-agent analysis with different models in different roles on miniF2F test set with sampling budget PAB@32. Results are averaged across 5 random seeds.}
\label{tab:multi_agent_results}
\begin{tabular}{@{}lllc@{}}
\toprule
\textbf{Generator} & \textbf{Decomposer} & \textbf{Critic} & \textbf{Success (\%)} \\
\midrule
Goedel & Kimina   & Kimina   & $81.4 \pm 0.4$ \\
Goedel & Kimina   & Goedel   & $82.2 \pm 0.5$ \\
Goedel & Goedel   & Kimina   & $82.7 \pm 0.4$ \\
Goedel & DeepSeek & DeepSeek & $80.9 \pm 0.3$ \\
Goedel & Goedel   & DeepSeek & $\mathbf{86.1 \pm 0.3}$ \\
Goedel & DeepSeek & Goedel   & $83.8 \pm 0.4$ \\
Goedel & Goedel   & Goedel   & $84.2 \pm 0.5$ $^{\dagger}$ \\
\bottomrule
\end{tabular}
\vspace{2pt}
\\
{\footnotesize $^{\dagger}$Baseline.}
\end{table}

\subsection{Effect of the Iteration--Branching Allocation}
\label{sec:n-s-ablation}

We study how the fixed proof-attempt budget should be distributed
between the number of MCTS iterations $N$ and the branching factor
$K$. We keep the number of complete proof attempts per child fixed at
$S=2$ and vary $N$ and $K$ such that
\[
N \times K \times S = 32
\]
for every configuration. Consequently, all settings use the same
maximum number of complete proof generations, decomposition calls, and
critic evaluations. Specifically, each configuration produces
$N K = 16$ decomposition candidates, $5 N K = 80$ critic evaluations,
and $N K S = 32$ complete proof attempts. The configurations therefore
differ primarily in how these calls are distributed across sequential
MCTS iterations.

\begin{table}[t]
\centering
\caption{Effect of the iteration--branching-factor allocation under a
fixed proof-attempt budget on MiniF2F with Goedel-Prover-V2-8B.
We fix $S=2$ proof attempts per child and vary the number of MCTS
iterations $N$ and children per expansion $K$, such that
$NKS=32$ for all configurations. Wall-clock time is reported relative
to the $N=1,K=16$ configuration. Results are averaged over five
random seeds.}
\label{tab:n-s-ablation}
\begin{tabular}{ccccc}
\toprule
$N$ & $K$ & $S$ & Success (\%) & Rel. Wall-Clock \\
\midrule
1  & 16 & 2 & $82.1 \pm 0.4$ & $1.0\times$ \\
2  & 8 & 2 & $82.3 \pm 0.3$ & $1.2\times$ \\
4$^{\dagger}$  & 4 &2  & $\mathbf{84.2 \pm 0.5}$ & $1.5\times$ \\
8 & 2 & 2  & $83.7 \pm 0.4$ & $2.1\times$ \\
16 & 1 & 2  & $84.3 \pm 0.2$ & $2.8\times$ \\
\bottomrule
\end{tabular}
\\[2pt]
\footnotesize $^{\dagger}$Configuration used in our main results.
\end{table}

As shown in Table~\ref{tab:n-s-ablation}, shallow configurations with
large branching factors are computationally efficient but achieve
lower proof success. The $N=1,K=16$ and $N=2,K=8$ configurations obtain
success rates of $82.1\%$ and $82.3\%$, respectively. In these settings,
most of the budget is allocated to generating a broad set of
decompositions, with limited opportunity to use intermediate rewards
to guide subsequent node selection.

Increasing the number of MCTS iterations to $N=4$ improves success to
$84.2\%$. Notably, the $N=1,K=16$ and $N=4,K=4$ configurations use the
same numbers of decomposition calls, critic evaluations, and complete
proof attempts. Their performance difference therefore cannot be
attributed to a larger component-call budget. Instead, the result
indicates that distributing node expansions across multiple MCTS
iterations, with intermediate reward backpropagation and subsequent
UCB-guided selection, is more effective than generating all
decomposition candidates in a single expansion.

Further increasing the number of iterations produces diminishing
returns. The $N=8,K=2$ configuration achieves $83.7\%$, while the
narrowest configuration, $N=16,K=1$, achieves $84.3\%$. The latter is
only $0.1$ percentage points above the $N=4,K=4$ configuration, while
requiring $2.8\times$ the wall-clock time of the $N=1,K=16$ setting,
compared with $1.5\times$ for $N=4,K=4$. This difference in success is
small relative to the variation across random seeds.

These results reveal an accuracy--latency trade-off. Broad, shallow
search reduces inference latency but provides insufficient iterative
refinement, whereas very narrow search increases sequential search
overhead without yielding a meaningful improvement in proof success.
We therefore use $N=4,K=4,S=2$ in the main experiments, as it provides
the best practical balance between proof success and wall-clock
efficiency under the fixed proof-attempt budget.

\paragraph{Reward Hacking Analysis.}
DeepSeek-Prover-V2-7B exhibits a substantially different pattern on PutnamBench.
Before exploit auditing, whole-proof sampling solves $13/659$ problems at
PAB@32 and $18/659$ at PAB@128, whereas our method solves $27/659$ and
$44/659$, respectively. However, screening the successful proofs using
previously reported exploit indicators identifies a substantial number
containing the \texttt{apply?} tactic and Cardinal-family lemmas
(\texttt{Cardinal.toNat}, \texttt{Cardinal.natCast\_inj}). These patterns were
identified in the DeepSeek-Prover-V2-7B technical report
\cite{ren2025deepseekproverv2advancingformalmathematical} as being associated
with a Lean 4.9.0 compiler/interface vulnerability in which \texttt{apply?}
may fail to emit an explicit \texttt{sorry} declaration under certain
conditions, producing proof declarations that compile despite depending on
\texttt{sorryAx}.

After excluding proofs confirmed by our audit to depend on \texttt{sorryAx},
the whole-proof solve counts decrease from $13/659$ to $9/659$ at PAB@32 and
from $18/659$ to $10/659$ at PAB@128. Thus, the audit removes 4 and 8
whole-proof solutions at the two budgets. For our method, the solve counts
decrease from $27/659$ to $16/659$ at PAB@32 and from $44/659$ to $25/659$ at
PAB@128, removing 11 and 19 solutions, respectively. Consequently, our method
produces 7 more exploit-dependent successful proofs than whole-proof sampling
at PAB@32 and 11 more at PAB@128.

% These absolute differences provide evidence that MCTS could amplify exploitative
% behaviour already present in the underlying prover. Under whole-proof
% sampling, exploit-containing proofs are generated independently, and a
% successful exploit does not affect subsequent samples. Under MCTS, in
% contrast, successful exploit-generating trajectories receive positive compiler
% rewards that are propagated through the tree. The UCB-guided policy can then
% allocate additional proof attempts to the corresponding nodes, causing later
% iterations to concentrate on trajectories that repeatedly produce compiling
% but axiom-dependent proofs rather than on genuine proof strategies.

Critically, successful exploitation is not observed on the physics benchmarks. On PhysLeandata, DeepSeek-Prover-V2-7B generates exploit-related patterns for only 1 of the 200
evaluated problems, and none of these attempts compile successfully. Likewise,
no exploit-dependent successful proof is identified on LeanPhysBench. These
results suggest that the prover lacks sufficient familiarity with the
graduate-level physics formalizations to construct the surrounding Lean
context required for the documented exploit pattern to succeed. We therefore
observe successful exploit-dependent proofs primarily on PutnamBench, where
the underlying prover already exhibits the documented behaviour. We report
both unaudited and audited results transparently in
Table~\ref{tab:putnam_results} and recommend kernel-level proof auditing for
models that exhibit anomalous increases in compiler-verified solve counts
under search-based inference.

\paragraph{Potential Exploit Identification.}
\label{sec:exploit_assess}
To investigate the generation of exploitative proof strategies, we adopt
a conservative multi-stage audit protocol. We first identify
\emph{potentially suspicious} proof attempts using the exploit patterns
reported in the DeepSeek-Prover-V2-7B technical report
\cite{ren2025deepseekproverv2advancingformalmathematical}. That work reports
that the 7B model occasionally exploits a Lean 4.9.0 compiler/interface bug on
PutnamBench through characteristic proof structures involving the
\texttt{apply?} tactic together with specific \texttt{Cardinal}-related lemmas,
including \texttt{Cardinal.toNat} and
\texttt{Cardinal.natCast\_inj}. We use these patterns only as an initial
screening mechanism; their lexical presence is never treated as sufficient
evidence that a proof is exploitative. Because we are not Lean 4 domain
experts, we rely on the indicators documented in prior work rather than
independently deriving new exploit signatures. Accordingly, our initial
screening can detect recurrences of the documented exploit family but cannot
by itself rule out undocumented exploit patterns.

For every successfully compiled proof, including proofs that do not contain
the previously reported lexical indicators, we append
\texttt{\#print axioms <theorem\_name>} and recompile the resulting code using
the same Lean evaluation environment. A proof is classified as a
\emph{potential exploit} only when the resulting theorem declaration depends
on \texttt{sorryAx}. For proofs that also contain a documented suspicious
construction, we additionally remove that construction and verify that the
proof no longer compiles. Thus, neither the presence of \texttt{apply?} nor
the use of Cardinal-family lemmas is independently sufficient for
classification. Only successful proofs whose resulting declarations depend
on \texttt{sorryAx} are removed from the audited solve counts. Since the
underlying Lean behaviour has already been documented in prior work
\cite{ren2025deepseekproverv2advancingformalmathematical}, we use the term
\emph{potential exploit} rather than claiming to have independently
established a new compiler vulnerability.

Across all experiments, successful exploit-dependent proofs are observed only
for DeepSeek-Prover-V2-7B on PutnamBench. Under whole-proof sampling, the audit
removes 4 successful proofs at PAB@32 and 8 at PAB@128. Under our MCTS
framework, it removes 11 successful proofs at PAB@32 and 19 at PAB@128. No
successful proof generated by Goedel-Prover-V2-8B or Kimina-Prover-Preview-Distill-7B is found to
depend on \texttt{sorryAx} under the same audit protocol. Although
DeepSeek-Prover-V2-7B occasionally generates proof attempts containing
exploit-related patterns on PhysLeandata, LeanPhysBench, and the MiniF2F test
set, none of these attempts both compile successfully and produce a theorem
declaration depending on \texttt{sorryAx}. They therefore do not contribute
to the reported solve counts. These observations suggest that our MCTS
framework does not introduce a new exploit strategy; instead, it could increase
the number of successful exploit-dependent proofs produced by a model that
already exhibits the documented behavior.

All our experiments, including the baselines, are conducted using Lean 4.15.0, Mathlib v4.15.0 (commit \texttt{9837ca9d}, dated 2025-01-05), and the Kimina Lean Server
2.0.0 container image. Although the underlying interface behavior was
originally documented under Lean 4.9.0
\cite{ren2025deepseekproverv2advancingformalmathematical}, we verify that proof
attempts exhibiting the documented \texttt{apply?}-based pattern continue to
produce theorem declarations depending on \texttt{sorryAx} under our pinned
Lean 4.15.0 and Mathlib environment.

To ensure that the audit is not limited to proofs matching the previously
reported lexical indicators, we execute \texttt{\#print axioms} on every
successfully compiled proof generated by all three prover models, across all
four benchmarks, methods, and proof-attempt budgets. Because
\texttt{\#print axioms} queries the dependencies of the resulting Lean
declaration rather than relying on the evaluation wrapper's surface-level
compilation status or source-level \texttt{sorry} scan, it directly determines
whether a successful declaration depends on kernel axioms such as
\texttt{sorryAx}. This exhaustive audit finds no \texttt{sorryAx} dependency
among successful proofs generated by Goedel-Prover-V2-8B or Kimina-Prover-Preview-Distill-7B
under any evaluated configuration.

The same audit procedure is applied uniformly to whole-proof sampling and
MCTS-generated proofs, enabling a controlled comparison between the two
inference procedures. We emphasize that this procedure verifies axiom-level
dependencies as reported by Lean; it does not constitute an independent
mathematical or semantic validation beyond the guarantees provided by Lean's
kernel and axiom-dependency tracking.

\section{Implementation details}

\paragraph{Reproduction protocol and comparability to published results}. All baselines in the paper are re-run in-house under a single pinned environment (Lean 4.15.0, Mathlib v4.15.0 commit 9837ca9d dated 2025-01-05, Kimina Lean Server 2.0.0), using the prompt templates in Section \ref{sec:prompts} and the decoding configurations in Section \ref{sec:inference}. Our whole-proof sampling numbers are consequently not directly comparable to the pass@k figures reported in the original model releases, and we do not present them as reproductions of those figures. The factors accounting for the difference are, first, published numbers are obtained under substantially older toolchain pins, and solve rates are highly sensitive to this choice: \cite{gu2025proofoptimizertraininglanguagemodels} measure the same Goedel-Prover-V2-32B checkpoint at 90\% pass@64 on MiniF2F and 86 solves on PutnamBench at pass@184 under Mathlib 4.9, but at 80\% and 75 solves respectively under Mathlib 4.19 a ten-point swing on MiniF2F attributable to the toolchain alone, with no change to the model. Second, published figures use model-specific prompt templates and, for Goedel-Prover-V2, an optional compiler-guided self-correction mode; we instead use one uniform prompt template and decoding configuration across all three provers and disable self-correction, so that every method under comparison differs only in its search procedure and not in its inference stack. 

% Third, the published figures are themselves not stable across sources: Goedel-Prover-V2-8B is reported at 84.6\% pass@32 on MiniF2F in the technical report \cite{lin2025goedelproverv2scalingformaltheorem} and at 83.0–83.3\% on the released model cards. We evaluate on the [state which] version of the MiniF2F test set throughout.

Our decision to pin a recent toolchain rather than reproduce the original evaluation environment is deliberate. The axiom-level audit in the main paper (Subsections: Reward Hacking Analysis, Potential Exploit Identification) shows that the Lean 4.9.0-era interface behaviour documented by \cite{ren2025deepseekproverv2advancingformalmathematical} admits proof declarations that compile successfully, pass the standard source-level sorry scan, and nonetheless depend on sorryAx. Compiler-verified solve counts obtained under that toolchain therefore cannot be taken at face value without a kernel-level audit, and re-deriving every baseline within a single pinned and exhaustively audited environment is what makes the internal comparisons in this paper meaningful. Every result reported in the main paper whole-proof sampling, BFS+CG, one-shot decomposition, our reproduction of Prover Agent, and our framework is internal to this environment, uses identical frozen checkpoints, and varies only the inference-time procedure.

\paragraph{Physlib library}  PhysLib is a Lean 4 support library introduced alongside LeanPhysBench in Lean4Physics \cite{li2025lean4physicscomprehensivereasoningframework}. It formalizes standard physical units, dimensional constants, and the associated definitions against which the benchmark's theorem statements are written. Following the evaluation protocol of \cite{li2025lean4physicscomprehensivereasoningframework}, we supply PhysLib as direct input context: the library is prepended to the theorem statement within the prompt, rather than being exposed through premise retrieval or an import mechanism.

We report LeanPhysBench results both with and without this context Table \ref{tab:leanphysbench_results}. The with-PhysLib condition matches the setting used in the original benchmark evaluation and is the primary comparison; the no-PhysLib condition measures how far a prover can proceed on physics formalizations without access to the domain-specific definitions its statements depend on. The protocol is applied identically to whole-proof sampling and to our framework at every budget, so the with/without contrast varies only the presence of library context.

\subsection{Inference configuration.}
\label{sec:inference}
The three agent roles use different decoding configurations
tailored to their respective objectives.

\begin{itemize}
\item \textbf{Generator.} Maximum generation length is
16,384 tokens with decoding temperature $\tau=0.7$.
This relatively higher temperature encourages diverse proof
attempts while remaining sufficiently stable for formal proof
generation.

\item \textbf{Decomposer.} Maximum generation length is
1,024 tokens. The decomposer uses the adaptive temperature
schedule described in Equation~\ref{eq:temp}, allowing broader
exploration during early search and progressively more focused
subgoal generation at deeper levels.

\item \textbf{Critic.} Maximum generation length is limited to
3 tokens with decoding temperature $\tau=0.3$. Since the critic
produces only scalar quality ratings rather than proofs, a low
temperature is used to obtain stable and reproducible evaluations.
\end{itemize}

\subsection{Prompt Configurations}
\label{sec:prompts}

We report the exact system prompts used for each of the three agent
roles below. The same prompt template is used across all three prover
models (Goedel-Prover-V2-8B, DeepSeek-Prover-V2-7B, Kimina-Prover-7B) and
all four benchmarks (MiniF2F, PutnamBench, PhysLeandata,
LeanPhysBench); only the theorem statement, decomposition trajectory,
and (where applicable) domain-specific library context are substituted
into the template placeholders at inference time.

\begin{promptbox}{Generator System Prompt}
\begin{lstlisting}[style=promptstyle]
You are an expert Lean4 theorem prover.
Given a theorem and a decomposition that breaks it into tractable steps, produce a complete Lean4 proof.

THEOREM:
{theorem}

DECOMPOSITION:
{decomposition}

INSTRUCTIONS:
- You may use a <think>...</think> block for internal reasoning
- After the thinking block, output ONLY the raw Lean4 proof, no markdown fences, no prose
- Start your Lean4 output directly with the theorem declaration or `by`
- Use `have` statements to encode each decomposition step as an intermediate goal
- The proof must typecheck and close all goals
- Do not output anything after the closing `done` or final tactic
\end{lstlisting}
\end{promptbox}

\begin{promptbox}{Critic System Prompt}
\begin{lstlisting}[style=promptstyle]
You are an expert Lean4 theorem proving evaluator.
You will be given:
1. A main Lean4 theorem to prove
2. A decomposition trajectory of prior steps (if any)
3. A candidate next decomposition step to evaluate

Rate the candidate step on how useful it is for making the main theorem provable.

Scoring criteria:
- Does it meaningfully reduce the proof burden of the main theorem? (0-25 pts)
- Is it consistent with and builds on the prior trajectory? (0-25 pts)
- Is it specific and immediately actionable in Lean4? (0-25 pts)
- Does it avoid redundancy with prior steps? (0-25 pts)

Score anchors:
- 0-20: Irrelevant, wrong direction, or harmful to the proof
- 21-40: Vaguely related but not actionable
- 41-60: Reasonable but generic or partially useful
- 61-80: Clearly useful and directly reduces proof burden
- 81-100: Optimal next step, precise and immediately actionable

Theorem: {theorem}
Trajectory: {trajectory}
Candidate step: {candidate_step}
Score:
\end{lstlisting}
\end{promptbox}

\begin{promptbox}{Decomposer Prompt}
\begin{lstlisting}[style=promptstyle]
You are an expert Lean4 proof strategist.
You will be given a theorem and optionally a list of decomposition steps already planned.
Your task is to output the single most useful NEXT proof step.

THEOREM:
{theorem}
{prior_steps_block}

INSTRUCTIONS:
- Output EXACTLY ONE next step, one sentence, plain English
- The step must be a concrete proof-structural move: an induction scheme, a case split,
  a key lemma to establish, a rewrite rule to apply, or an algebraic transformation
- It must directly reduce what remains to be proven
- Do NOT write Lean4 code or pseudocode
- Do NOT explain your reasoning
- Do NOT repeat prior steps
- Do NOT state what the final answer is

Valid examples:
  "Apply induction on n, handling the base case n=0 and step case separately."
  "Rewrite the goal using the identity a^2 - b^2 = (a-b)(a+b)."
  "Case split on whether k is even or odd."
  "Establish the auxiliary lemma that f(n) is monotone increasing, then apply it to bound the sum."
  "Reduce to showing divisibility by expressing the left side as a telescoping product."

Next step:
\end{lstlisting}
\end{promptbox}

\subsection{Example: A Potential Compiler Exploit}
\label{sec:exploit-example}

Listing~\ref{lst:exploit-example} shows a representative
\texttt{Agent (critic+reward)} proof attempt for
\texttt{putnam\_1997\_b5} generated by DeepSeek-Prover-V2-7B that was
flagged by our audit protocol ((Subsections: Reward Hacking Analysis, Potential Exploit Identification), stated in the main paper). The proof
compiles successfully under the Kimina Lean server and superficially
follows a standard structure: a base case (\texttt{h\_base}), an
inductive step (\texttt{h\_inductive}), and a chain of trivial
\texttt{exact} forwarding (\texttt{h\_main}, \texttt{h\_final}) that
gives the appearance of a completed induction. Critically, however,
the inductive step's proof obligation is discharged not through valid
tactics but through a cascade of \texttt{try} blocks that falls back
to the \texttt{apply?} tactic (highlighted, line~\ref{line:applyq})
when \texttt{omega} and \texttt{simp\_all} fail to close the goal.

Following our audit protocol, we append an axiom check to the
compiled proof:
\begin{lstlisting}[style=promptstyle]
#print axioms putnam_1997_b5
\end{lstlisting}
which returns:
\begin{lstlisting}[style=promptstyle]
'putnam_1997_b5' depends on axioms: [propext, sorryAx, Classical.choice, Quot.sound]
\end{lstlisting}
The presence of \texttt{sorryAx} confirms that, despite compiling
without an explicit \texttt{sorry} token in the source and despite
passing the Kimina server's \texttt{sorry}-token scan, the proof does
not in fact close all goals \texttt{apply?} silently discharges an
unproved obligation via the Lean~4.9.0 interface bug documented in
\citet{ren2025deepseekproverv2advancingformalmathematical}, rather than genuinely establishing the
inductive step. Removing the flagged \texttt{apply?} branch causes the
proof to fail to compile, satisfying the second criterion of our audit
protocol ((Subsections: Reward Hacking Analysis, Potential Exploit Identification), stated in the main paper)) and confirming classification as a
potential exploit.

% Full listing for supplementary document
\begin{lstlisting}[style=promptstyle, caption={A DeepSeek-Prover-V2-7B proof attempt for \texttt{putnam\_1997\_b5} flagged as a potential exploit. Line~\ref{line:applyq} shows the fallback \texttt{apply?} tactic responsible for the \texttt{sorryAx} dependency.}, label={lst:exploit-example}, escapeinside={(*}{*)}]
import Mathlib
def tetration : Nat -> Nat -> Nat
  | _, 0 => 1
  | b, (m + 1) => b^(tetration b m)

theorem putnam_1997_b5
(n : Nat)
(hn : n >= 2)
: tetration 2 n === tetration 2 (n-1) [MOD n] := by
  have h_base : n = 2 -> tetration 2 n === tetration 2 (n-1) [MOD n] := by
    intro hn2
    rw [hn2]
    norm_num [tetration, Nat.ModEq]
    <;> decide

  have h_inductive : forall k >= 2, tetration 2 k === tetration 2 (k-1) [MOD k]
      -> tetration 2 (k+1) === tetration 2 k [MOD (k+1)] := by
    intro k hk h
    have h1 : tetration 2 (k + 1) = 2 ^ (tetration 2 k) := by
      rw [tetration]
      <;> simp [Nat.add_comm, Nat.add_assoc, Nat.add_left_comm]
      <;> induction k <;> simp_all [tetration, Nat.pow_succ]
      <;> ring_nf at * <;> omega
    have h2 : tetration 2 k = 2 ^ (tetration 2 (k - 1)) := by
      cases k with
      | zero => contradiction
      | succ k =>
        cases k with
        | zero => contradiction
        | succ k =>
          simp_all [tetration]
          <;> simp_all [tetration, Nat.pow_succ]
          <;> ring_nf at * <;> omega
    rw [h1]; rw [h2]
    have h3 : (2:Nat) ^ (2 ^ (tetration 2 (k - 1))) = 2 ^ (2 ^ (tetration 2 (k - 1))) := rfl
    have h4 : (2:Nat) ^ (2 ^ (tetration 2 (k - 1))) = 2 ^ (2 ^ (tetration 2 (k - 1))) := rfl
    simp_all [Nat.ModEq, Nat.ModEq]
    <;> (try omega) <;>
    (try { cases k with
           | zero => contradiction
           | succ k =>
             cases k with
             | zero => contradiction
             | succ k => simp_all [tetration, Nat.pow_succ, Nat.ModEq] <;> omega }) <;>
    (try { simp_all [tetration, Nat.pow_succ, Nat.ModEq] <;> omega }) <;>
    (try { apply? }) (*\label{line:applyq}*)
    <;> omega

  have h_main : forall k >= 2, tetration 2 k === tetration 2 (k-1) [MOD k] := by
    intro k hk
    induction' hk with k hk ih
    . norm_num [tetration, Nat.ModEq]
    . cases k with
      | zero => contradiction
      | succ k =>
        cases k with
        | zero => norm_num [tetration, Nat.ModEq] at *
        | succ k =>
          have h1 := h_inductive (k + 2) (by omega) ih
          simp_all [tetration, Nat.ModEq]
          <;> omega

  exact h_main n hn
\end{lstlisting}

\section{Datasets}

We evaluate our framework on four publicly available Lean 4 theorem-proving benchmarks spanning competition mathematics and college- to graduate-level physics.

\textbf{MiniF2F} \cite{zheng2022minif2fcrosssystembenchmarkformal} is a benchmark consisting of 244 test problems formalized in Lean 4. The problems are derived from international mathematics competitions, including the International Mathematical Olympiad (IMO), the American Invitational Mathematics Examination (AIME), and related contests. The benchmark covers a diverse range of mathematical domains, including algebra, number theory, geometry, inequalities, combinatorics, induction, and analysis, making it a standard benchmark for evaluating formal theorem provers.

\textbf{PutnamBench} \cite{tsoukalas2024putnambenchevaluatingneuraltheoremprovers} is a collection of formally verified problems compiled from the William Lowell Putnam Mathematical Competition. It contains challenging undergraduate-level mathematical theorems spanning algebra, analysis, geometry, combinatorics, number theory, and linear algebra, providing a rigorous benchmark for assessing mathematical reasoning and proof-generation capabilities.

\textbf{LeanPhysBench} \cite{li2025lean4physicscomprehensivereasoningframework} is a benchmark of formally verified college-level physics problems written in Lean 4. The dataset covers multiple introductory and intermediate physics domains, including mechanics, oscillations and waves, electromagnetism, thermodynamics, optics, and related topics. We evaluate both with and without the accompanying PhysLib context following the benchmark protocol.

\textbf{PhysLeanData} \cite{zhang2026physproveradvancingautomatictheorem} is a large-scale Lean 4 benchmark containing graduate-level theoretical physics formalizations. The benchmark includes problems from advanced domains such as classical particle mechanics, classical string theory, relativity, quantum mechanics, quantum field theory, and related areas, providing an out-of-distribution evaluation setting for assessing the generalization capability of theorem-proving systems.

\section{Conclusion and Limitations}
\label{sec:conclusion}

We present a three-role MCTS framework that treats the Lean 4 compiler purely as a reward oracle, this prevents compiler-error content from accumulating in the generation context as search depth increases. This keeps compiler-feedback overhead constant with search depth. Under matched proof-attempt budgets, our method outperforms whole-proof sampling across four benchmarks spanning competition mathematics and graduate-level theoretical physics, reaching 87.1\% on MiniF2F with Goedel-Prover-V2-8B at PAB@256, while consuming 32.8\% fewer total inference tokens on average than the whole-proof baseline. Ablations show that both the temperature-decayed decomposer and the distribution of a fixed budget across multiple MCTS iterations, rather than a single wide expansion, are responsible for these gains. We also present an exhaustive axiom-level audit. We ran \texttt{\#print axioms} on every successfully compiled proof produced by all three prover models across all four benchmarks, methods, and budgets not only on proofs matching previously reported lexical indicators. This audit shows that a substantial fraction of DeepSeek-Prover-V2-7B's successes on PutnamBench depend on sorryAx despite compiling cleanly and passing the standard source-level sorry scan: 4 of 13 and 8 of 18 whole-proof successes at PAB@32 and PAB@128, and 11 of 27 and 19 of 44 under our framework. The underlying apply? interface behaviour was documented for Lean 4.9.0 by Ren et al. (2025); we confirm it persists under our pinned Lean 4.15.0 and Mathlib environment.

% and no exploit-dependent success appears on the out-of-distribution physics benchmarks, where the prover appears unable to construct the surrounding Lean context the exploit pattern requires.

\appendix

% Finally, the fixed tree
% structure ($K{=}4$, $D_{\max}{=}4$) and budget scaling through $S$ alone
% leaves open the question of optimal resource allocation across branching
% factor, depth, and rollout count for different problem difficulty levels ---
% an adaptive allocation strategy informed by intermediate reward signals
% could further improve budget efficiency beyond what our current fixed
% structure achieves.

% Using the \centering command instead of \begin{center} ... \end{center} will save space
% Positioning your figure at the top of the page will save space and make the paper more readable
% Using 0.95\columnwidth in conjunction with the

\bibliography{ref}


\end{document}